\documentclass[runningheads]{llncs}
\usepackage[T1]{fontenc}
\usepackage{graphicx}
\usepackage{booktabs}
\usepackage{multirow}
\usepackage{rotating}
\usepackage{float}
\usepackage{amsmath}
\usepackage{mathtools}
\usepackage{xcolor}
\usepackage{hyperref}
\usepackage{longtable}
\usepackage{colortbl}
\usepackage{siunitx}
\usepackage{graphicx}
\usepackage[export]{adjustbox}
\usepackage{diagbox}

\usepackage[table]{xcolor}

\begin{document}
\title{Benchmarking SAM2-based Trackers on FMOX}

%
%
\author{Senem Aktas\orcidID{0000-0002-3996-2771} \and Charles Markham\orcidID{0000-0003-2447-2611} \and John McDonald\orcidID{0000-0001-9225-673X} \and Rozenn Dahyot\orcidID{0000-0003-0983-3052}}
\authorrunning{Aktas et al.}
\institute{Department of Computer Science, Maynooth University, Ireland}

\maketitle              
\begin{abstract}
Several object tracking pipelines extending Segment Anything Model 2 (SAM2) have been proposed in the past year, where the approach is to follow and segment the object from a single exemplar template  provided by the user on a initialization frame. 
We propose to benchmark these high performing trackers (SAM2, EfficientTAM, DAM4SAM and SAMURAI) on  datasets containing fast moving objects (FMO) specifically designed to be challenging for tracking approaches. The goal is to understand better current limitations in state-of-the-art trackers by providing more detailed insights on the behavior of these trackers. We show that overall the trackers DAM4SAM and SAMURAI perform well on more challenging sequences.

\keywords{Segment anything model  \and Fast moving object \and Tracking.}
\end{abstract}

\section{Introduction}

Video Object Tracking (VOT) and Video Object Segmentation (VOS) both  aim to follow the object throughout a video or image sequence. VOT, focuses on locating and following the position of a target object, typically outputting a bounding box without providing detailed shape or pixel-level information. While VOS, seeks to identify and segment the object at the pixel level, producing a mask in each frame, thereby capturing its shape and boundaries \cite{ding2025mosev2}.

SAM2 is one of the state-of-the-art (SOTA)  VOS/VOT methods that uses object positions given as points, bounding boxes, or masks in any frame to initialize tracking \cite{ding2025mosev2}. Various extentions of SAM2 has been proposed for specific purposes; for instance DAM4SAM \cite{videnovic2025distractor} for handling distractors, SAMURAI \cite{yang2024samurai} managing fast motions, and EfficientTAM \cite{xiong2024efficient} for improving efficiency across various platforms (cf. Section \ref{section:sam2_trackers_explained}).

Recent SAM2-based works proposed the DiDi dataset \cite{videnovic2025distractor} to address distractors, and the Mosev2 dataset \cite{ding2025mosev2} to handle various challenging cases. The study in \cite{ding2025mosev2} utilized SAM2 and its variants, including DAM4SAM \cite{videnovic2025distractor} and SAMURAI \cite{yang2024samurai}. Similarly, Aktas et al \cite{Aktas2025} introduced FMOX, a JSON format designed for challenging Fast Moving Object (FMO) datasets, and extended the ground truth annotations to include object size categorization. FMOX \cite{Aktas2025} has been used to evaluate the SAM-based tracker EfficientTAM, demonstrating its performance compared to FMO-specific pipelines \cite{Rozumnyi2021IJCV,kotera2020restoration,rozumnyi2017world,kotera2019intra,rozumnyi2020sub} using the Trajectory Intersection over Union (TIoU) metric.

In this paper, we extend the benchmarking of SAM2 and its variants DAM4SAM, and SAMURAI, alongside EfficientTAM on FMOX dataset through the use of more standard performance metrics: the Mean Intersection over Union (mIoU) and the Dice Score metrics. These generalized metrics were prioritized to facilitate a broader comparison against the wider state-of-the-art literature. The results indicate that DAM4SAM consistently outperforms the other trackers on FMO datasets, aligning with similar observations from Mosev2 \cite{ding2025mosev2}. 
On the other hand, EfficientTAM shows comparatively lower performance among the SAM2-based trackers examined.
In Section~\ref{section:background}, we provide a brief background on the datasets and SAM2-based trackers. Section \ref{section:methodology} presents our methodology for benchmarking   along with the tracker initialisation process and performance analysis. In Section \ref{sec:Result}, we present and discuss our findings, and finally, Section \ref{section:conclusion} summarises the conclusions drawn from this study.

\section{Background} \label{section:background}

\subsection{Benchmark datasets for object tracking}

Despite the abundance of video and image benchmarks, many challenges remain unaddressed in object tracking  which limit their ability to generalize to complex real-world scenarios \cite{ding2025mosev2}. The recently proposed MOSEv2 dataset (coMplex video Object SEgmentation) \cite{ding2025mosev2} addresses several of these difficulties by including videos with complex scenes featuring object disappearance and reappearance, heavy occlusions, crowded areas, small objects, poor lighting, and camouflage. Ding et al \cite{ding2025mosev2} use MOSEv2 dataset to evaluate trackers such as SAM2 \cite{ravi2024sam}, SAMURAI \cite{yang2024samurai}, and DAM4SAM \cite{videnovic2025distractor}. 

Even though some challenges remain in standard benchmarks, the overall results reported often fail to reflect these difficulties, as many trackers do not effectively capture or address them. This leads to inflated performance scores that mask the true complexity of real-world tracking scenarios \cite{videnovic2025distractor}. Addressing this gap, DAM4SAM \cite{videnovic2025distractor} focused on distractors and occlusions by carefully selecting validation and test sequences from major benchmarks including LaSOT \cite{fan2019lasot} and GOT-10k \cite{cui2022mixformer}, forming the DiDi dataset to enable more rigorous evaluation.

A similar argument can be made for Fast Moving Objects (FMOs), which represent a significant yet often overlooked challenge in tracking due to their high speed and motion-induced blur. These characteristics limit the effectiveness of many current trackers and are not adequately captured by standard benchmarks. This is particularly important given that the FMO problem is essential for advancing tracking performance in practical applications, such as sports analysis, autonomous driving, and robotics.

\begin{figure}[!h]
\begin{center}
\includegraphics[width=.8\linewidth]{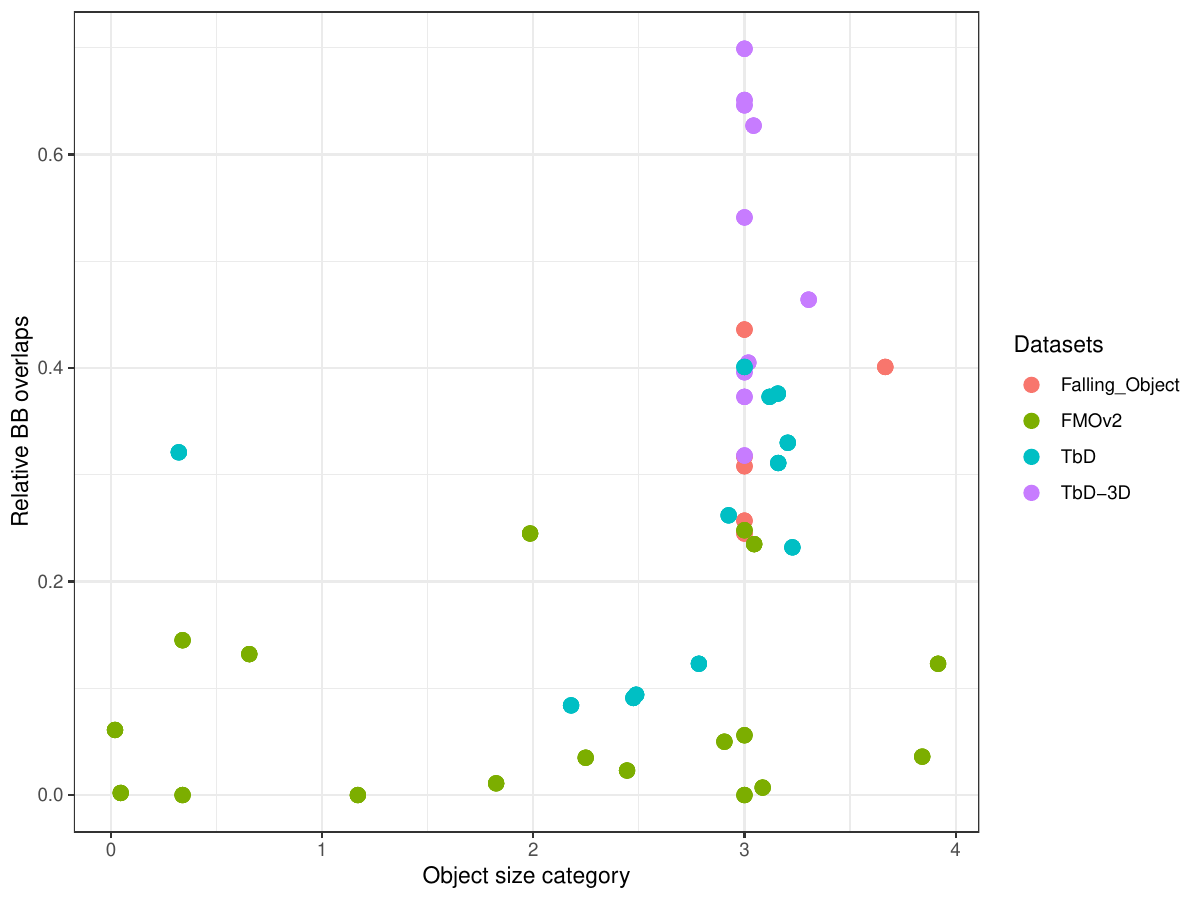}
\end{center}
\vspace{-.5cm}
\caption{FMOX regroups 4 datasets, with a total 46 sequences used for our benchmark. Using ground truth information, object size is divided into 5 categories (0 for \textit{Extremely tiny} up to 4 for \textit{large} \cite{Aktas2025}) with a mean object size  is computed for each sequence (reported on the $x$-axis). To capture displacement between two successive frames, the mean IoU between two successive ground truth bounding boxes is also computed for each sequence (reported on the $y$-axis - BB represents bounding boxes). In contrast to \texttt{Falling Object} and \texttt{TbD-3D}, both \texttt{FMOv2} and \texttt{TbD} datasets are more challenging  having  smaller objects with smaller overlapping successive  bounding boxes.}
\label{fig:datasets}
\end{figure}

We focus here on datasets specifically designed for evaluating FMOs, including Falling Object (6 sequences, \cite{kotera2020restoration}),  TbD (12 sequences, \cite{kotera2019intra}),  TbD-3D (10 sequences, \cite{rozumnyi2020sub}) and FMOv2 for which 18 sequences
are used here from the 19 available (\cite{rozumnyi2017world}, the sequence \texttt{more\_balls}  with multiple objects has been excluded from our analysis  due to its multi-instance objects lacking unique IDs \cite{Aktas2025}). 
Each dataset offers unique features: for instance, TbD-3D extends the challenge by incorporating 3D object motion and appearance changes. While FMOv2 is designed to be more challenging, with high object displacements and almost no bounding box overlap (IoU) between consecutive frames.
These datasets (collectively named the FMOX dataset) have been recently augmented with ground truth JSON files to enable straightforward and easy-to-use benchmarking of trackers \cite{Aktas2025}. FMO datasets not only include fast-moving objects but also small objects where the FMOX description can be used to focus on specific object size categories \cite{Aktas2025} such as small objects which are challenging for tracking (c.f. Fig. \ref{fig:datasets}). We provide benchmark results on the FMOX dataset in Section \ref{sec:Result} for showcasing the capabilities of these various trackers. 

\subsection{Segment Anything Model 2 (SAM2) Based Trackers}\label{section:sam2_trackers_explained}

The Efficient Track Anything Model (EfficientTAM) \cite{xiong2024efficient}, Distractor-Aware Memory for SAM2 (DAM4SAM) \cite{videnovic2025distractor}, and SAM-based
Unified and Robust zero-shot visual tracker with motion-Aware Instance-level memory (SAMURAI) \cite{yang2024samurai} trackers are SAM2 \cite{ravi2024sam} based trackers. A primary reason for choosing SAM2 \cite{ravi2024sam} based trackers is that they do not require training or retraining. Each tracker offers distinct advantages in memory management, performance and adaptation for visual object tracking. Specifically, DAM4SAM is selected to avoid distractors near the target object, particularly in scenarios involving multiple instances of the target object. Meanwhile, SAMURAI is chosen for its integrated motion modeling and motion-aware instance-level memory, which enhance performance in crowded scenes with fast-moving or self-occluding objects, where SAM2 encounters challenges. EfficientTAM is chosen for its ability to deliver accelerated performance and reduced computational costs, making it ideal for efficient video object segmentation across various platforms. 

\subsubsection{SAM2.} 

The image encoder and memory mechanism are essential components of SAM2. The image encoder is responsible for extracting features from frames, while the memory mechanism stores the past \(n\) frames to facilitate the segmentation of new frames \cite{xiong2024efficient}. This memory mechanism consists of 7 slots for storing 7 frames, with the first slot reserved for the initialized frame. The remaining 6 slots are updated each time a new frame arrives, following a first-in, first-out (FIFO) queue method \cite{xiong2024efficient,videnovic2025distractor,yang2024samurai}. SAM2 generates three output masks and selects the one with the highest predicted Intersection over Union (IoU). However, DAM4SAM \cite{videnovic2025distractor} noted that simple output masks selection often leads to the inclusion of distractors from previous frames before a tracking failure occurs due to the accumulation of misleading information. Additionally, this straightforward approach can create further issues in crowded scenes where target and background objects have similar appearances. Simply relying on the previous \(n\) frames can also result in the storage of misleading features during occlusion \cite{yang2024samurai}.

\subsubsection{EfficientTAM.}

EfficientTAM offers a lightweight version of SAM2 to reduce the high computational complexity of the image encoder and memory module, particularly for video object segmentation on mobile devices. Unlike the original SAM2, EfficientTAM adopts a lightweight Vision Transformer (ViT) image encoder for improved efficiency. In the memory mechanism, tokens are small pieces of information that the model uses to remember different parts of an image or data. Two adjacent tokens are similar, with a small difference between them, defined by a constant that specifies the acceptable level of similarity. To enhance memory efficiency, EfficientTAM avoids storing multiple nearly identical memory tokens for similar parts of an image or data. Instead, it consolidates these similar tokens into a single representative token. This means that rather than keeping separate tokens for each similar part, the model creates one token that captures the essence of all those similar parts. As a result, the overall set of tokens becomes a coarser representation of the original, meaning it retains the same total number of tokens but simplifies the information they represent. By doing this, EfficientTAM can process information using fewer unique tokens. This not only speeds up the calculations but also reduces the amount of memory required, making the model more efficient.

\subsubsection{DAM4SAM.}

Distractors are elements within the visual field that complicate the tracking of a target object. These can be categorized into two types: external distractors, which are nearby objects that share visual similarities with the target such as an independent instance of the target object, and internal distractors, which are similar regions found on the target itself when only a portion of it is being tracked. The challenge posed by external distractors is particularly pronounced when the target exits and subsequently re-enters the field of view, as these similar-looking objects can lead to confusion in accurately identifying the target. To reduce to distractors failures, the FIFO memory update mechanism of SAM2 has been replaced with a Distractor-Aware Memory (DAM) management strategy. It divides the memory into Recent Appearance Memory (RAM) and Distractor Resolving Memory (DRM), utilizing a new memory management protocol for updates. The 3 slots in the RAM are updated every 5 frames with the FIFO mechanism in case the target object already exists. The DRM, which accounts for the remaining four slots, fixes the first slot in the initialization frame as SAM2. 

\subsubsection{SAMURAI.} SAMURAI adapts the SAM2 model to handle distractors and incorporates motion cues for improved memory management. A Kalman filter-based motion modeling is integrated to manage fast-moving and occluded objects in crowded scenes. In addition to the mask affinity score and object occurrence score, the output of the motion modeling, referred to as the motion score, is used to select frames for memory, rather than relying on the n-previous frames as SAM2 does. Instead of relying on a fixed window of frames, a dynamic frame selection process referred to as "Motion-Aware Instance-Level Memory" selectively chooses only the most reliable frames from a sequence to update the memory. A frame is considered a valid candidate for memory if it achieves a good affinity and motion score. If the remaining memory slots have not been filled, the tracked object is considered to be occluded or has disappeared, and frames are filled accordingly. Conversely, frames are discarded if they have a poor affinity or motion score. This approach ensures that the memory is composed of high-confidence frames.

\section{Method for benchmarking} 
\label{section:methodology}

\subsection{Pretrained models for benchmarking}

All trackers have been initialized via ground-truth bounding boxes and evaluated under their default model configurations: SAM2 and DAM4SAM with the \texttt{SAM2.1 Hiera Large (Hiera-L)} model, SAMURAI with the \texttt{SAM2.1 Hiera Base Plus (Hiera-B+)} model, and EfficientTAM with the \texttt{efficienttam\_s} model.

\subsection{Tracker Initialisation}

As observed in previous recent works \cite{Aktas2025,ding2025mosev2}, initialization of trackers with bounding boxes performs better than using points to provide an exemplar template to target in the following frames. We have chosen here to initialize all trackers with the first bounding box of the FMOX-labelled object when it occurs in each sequence.
Of note,  sometimes this first bounding box is not in the first frame of the sequence but occurs in a later frame.

\subsection{Tracker scoring}

The performance metrics Dice and IoU have been chosen here to compare  the tracker predicted bounding boxes with the ground truth ones.
These metrics are  computed on each frame for which FMOX provides a ground truth bounding box. 
Frames without a ground truth bounding box, for instance when the object has disappeared from view, are not taken into account in the computation of the sequence mean IoU (mIoU)  and mean Dice (mDice).
IoU, Dice, and their respective means computed over sequence are values between 0 (for object missed or not tracked) and 1 (for perfect detection and tracking). 

The first frame used for initialization of the tracker is omitted from performance calculations because its object location is provided by the FMOX  ground truth. 
Furthermore, for frames where the tracker fails to predict a bounding box, the IoU and Dice scores are set to zero, representing the worst possible scores for these frames. 
These zeros are included as part of the scores computed for each sequence in FMOX (mIoU and  mDice). 

\subsection{FMOX for benchmarking trackers}

To evaluate the performance of each tracker, we utilise the 46 sequences of the FMOX dataset, none of which were used during the training process. 
Hence we ensure that no data leakage has occurred between the models and the evaluation dataset.

\section{Results and discussion} 
\label{sec:Result}

\subsubsection{Quantitative results.} Table \ref{tab:performance} provides  the minimum, maximum, mean, and median values of mIoU and mDice metrics computed for  FMOX dataset for each of the trackers. Both metrics concur in finding the best overall performance with DAM4SAM using both the mean and median (equivalent to a robust mean) computed with all sequences in FMOX.
In contrast, EfficientTAM  has the worst performance of the four trackers tested.

\begin{table}[h!]
    \centering  
    \caption{Overall results obtained on  the 46 sequences in the FMOX dataset. Both the mean and median, computed with the  mDICE and mIoU for each sequence in FMOX, show DAM4SAM performing best (in bold font and an asterisk (*)). However the zeros scores observed  for the minimum highlight that trackers completely fails for some sequences, while the maximum scores show that sometimes SAM2 outperforms other tracker for some sequences. Values range from [0, 1] (0 = bad, 1 = good).}
    \label{tab:performance}

    \resizebox{\textwidth}{!}{%
        \begin{tabular}{p{.2\textwidth}p{.2\textwidth}p{.2\textwidth}p{.2\textwidth}p{.2\textwidth}}
            \toprule
            \textbf{mIoU}$(\uparrow)$& \textbf{SAM2} & \textbf{EfficientTAM} & \textbf{DAM4SAM} & \textbf{SAMURAI} \\
            \midrule
            MIN & 0.000 & 0.000 & 0.000 & 0.001 \\
            MAX & 0.928 & 0.799 & 0.819 & 0.925 \\
            MEDIAN & 0.591 & 0.548 & \textbf{0.605*} & 0.596 \\
            MEAN & 0.461 & 0.438 & \textbf{0.505*} & 0.488 \\
            \bottomrule
        \end{tabular}
    }%
    
    \vspace{5pt} 

    \resizebox{1\textwidth}{!}{%
        \begin{tabular}{p{.2\textwidth}p{.2\textwidth}p{.2\textwidth}p{.2\textwidth}p{.2\textwidth}}
            \toprule
            \textbf{mDice} $(\uparrow)$ & \textbf{SAM2} & \textbf{EfficientTAM} & \textbf{DAM4SAM} & \textbf{SAMURAI} \\
            \midrule
            MIN & 0.000 & 0.000 & 0.000 & 0.002 \\
            MAX & 0.962 & 0.885 & 0.899 & 0.961 \\
            MEDIAN & 0.699 & 0.684 & \textbf{0.744*} & 0.736 \\
            MEAN & 0.545 & 0.520 & \textbf{0.600*} & 0.579 \\
            \bottomrule
        \end{tabular}
    }
\end{table}

\begin{figure}[!h]
\begin{center}
\begin{tabular}{cc} 
  \includegraphics[height=5cm,trim=5 0 90 0, clip]{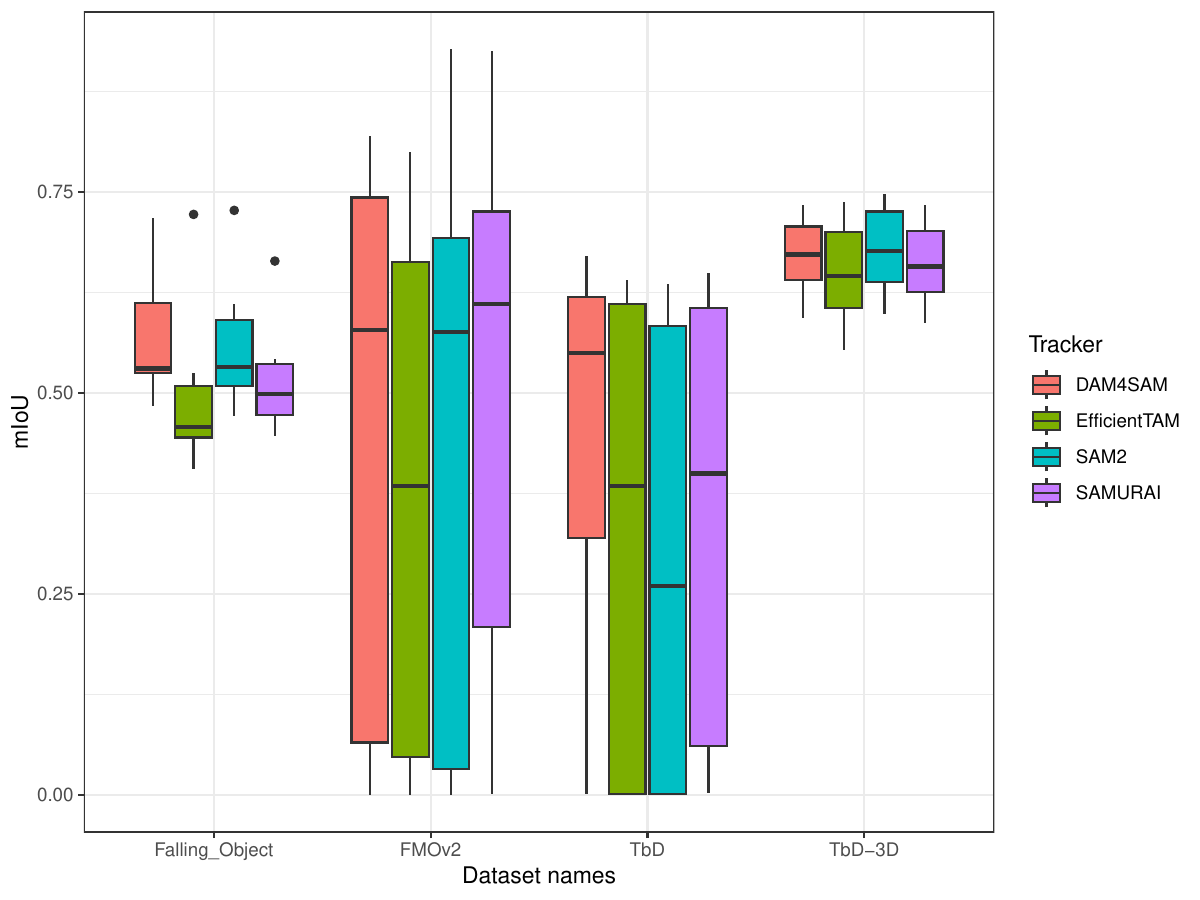} &
  \includegraphics[height=5cm]{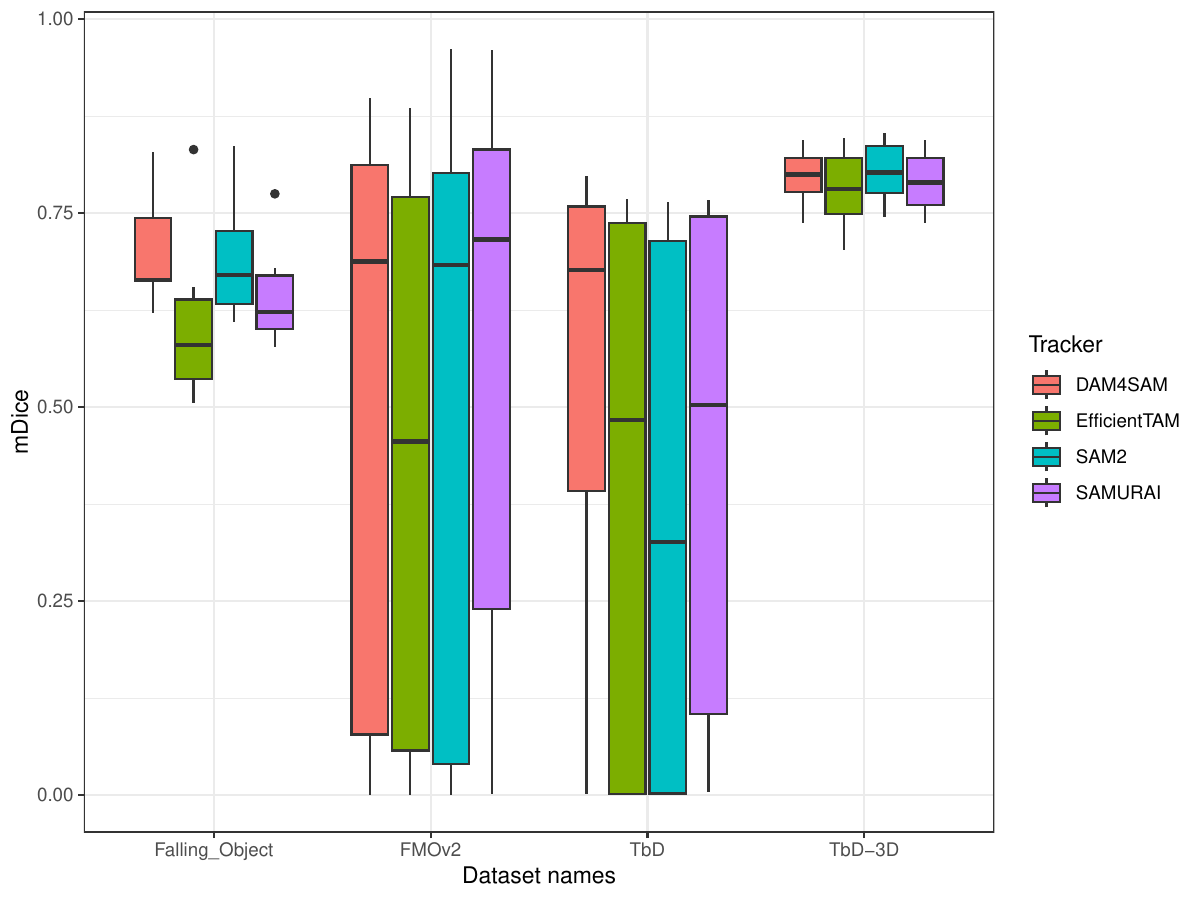} \\
\end{tabular}
\end{center}
\vspace{-0.5cm}
\caption{Box plots for mIoU and mDice results on each of the 4 datasets (reported on the x-axis) included in FMOX. Both  \textit{FMOv2} and \textit{TbD} are more challenging as objects tracked are of smaller sizes with often non overlapping ground truth bounding boxes between frames $n$ and $n+1$ (cf. Fig. \ref{fig:datasets}). The  low minima (=0) highlight the challenge presented by some sequences in these datasets for the trackers tested. }
\label{fig:performance:alldatasets}
\end{figure}

In Table \ref{table:perfromance_ranking}, we present the performance ranking of each tracker across each FMO datasets with box plots presented in Figure \ref {fig:performance:alldatasets}.
Our findings align with those reported in Mosev2 \cite{ding2025mosev2}, as DAM4SAM consistently outperforms other trackers on the FMO datasets, achieving the highest median and average mIoU and mDice scores. This indicates that DAM4SAM delivers both accurate and stable tracking performance across diverse datasets. EfficientTAM ranks lowest overall, with the poorest median and mean scores and frequent missed detections, underscoring its limitations in these challenging scenarios. SAMURAI and SAM2 demonstrates moderate performance, generally outperforming EfficientTAM. The results indicate that a tracker may perform strongly on some subsequences, while others exhibit poor or no performance. As highlighted in \cite{Aktas2025}, initializing trackers with highly motion-blurred frames can adversely affect their performance. Although EfficientTAM has been shown to perform competitively  to pipelines dedicated to track fast moving objects  \cite{Aktas2025}, its primary design focus is on reducing the computational cost of SAM2, making it more suitable for deployment across various platforms. 

\begin{table}[!h]
\centering
\caption{Model performance rankings per dataset in FMOX based on IoU and Dice Score. Only \textit{FMOv2} and \textit{TbD} have sequences with object size extremely tiny to small as per classification provided in FMOX JSON \cite{Aktas2025}. In addition, ground truth bounding boxes rarely overlap between frames $n$ and $n+1$ in the \textit{FMOv2} and \textit{TbD} datasets.  }
\begin{tabular}{p{4cm}ll}
\toprule
\textbf{Datasets in FMOX} & \textbf{Ranking (best to worst)} \\
\midrule
\textit{Falling Object} & DAM4SAM $>$ SAM2 $>$ SAMURAI $>$ EfficientTAM \\ [0.5em]
\textit{TbD-3D} & SAM2 $>$ DAM4SAM $>$ SAMURAI $>$ EfficientTAM \\ [0.5em]
\textit{FMOv2} & SAMURAI $>$ DAM4SAM $>$ SAM2 $>$ EfficientTAM \\ [0.5em]
\textit{TbD} & DAM4SAM $>$ SAMURAI $>$ EfficientTAM $>$ SAM2 \\ 
\bottomrule
\end{tabular}\label{table:perfromance_ranking}
\end{table}

\subsubsection{Compute costs.} The computational workload for each of the four trackers was processed using an NVIDIA GeForce RTX 4090 GPU. All experiments are conducted on a workstation equipped with a 13th Gen Intel Core i9 processor, 64 GB of RAM, CUDA 12.4.1, and Ubuntu 20.04.6 running on Windows Subsystem for Linux (WSL). Table \ref{tab:execution_times} reports computation times: as expected EfficientTAM is the fastest. 
\begin{table}[!h]
\begin{center}
\caption{Execution times (in seconds $\downarrow$) for the tested trackers across the 4 FMO datasets in FMOX. 
EfficientTAM offers the lowest computational overhead, ranking as the fastest tracker on all datasets. SAMURAI is the second fastest tracker but also has good accuracy in contrast to EfficientTAM (cf. Tab. \ref{tab:performance}).} 
\label{tab:execution_times}
\footnotesize
\begin{tabular}{|p{2.5cm}|w{c}{2cm}|w{c}{2cm}|w{c}{2cm}|w{c}{2cm}|}
 \hline
 \diagbox{Dataset}{Tracker}  & DAM4SAM & SAMURAI & EfficientTAM & SAM2 \\ 
 \hline
 Falling Object & 67.16  & 27.45 & \textbf{24.64} & 51.29  \\  \hline
 FMOv2 & 1316.92  & 515.73 & \textbf{410.07} &  890.80 \\  \hline
 TbD & 261.42 & 93.68 & \textbf{67.31} & 394.44 \\   \hline
 TbD-3D & 168.44 & 56.32  & \textbf{39.44}  &  203.92  \\
\hline
 \end{tabular}
 \normalsize

 \end{center}
\end{table}

\subsubsection{IoU per frames.} 
\begin{figure}[!h]
\begin{tabular}{c}
\includegraphics[width=\linewidth,trim=1cm .5cm 1.5cm 1cm,clip]{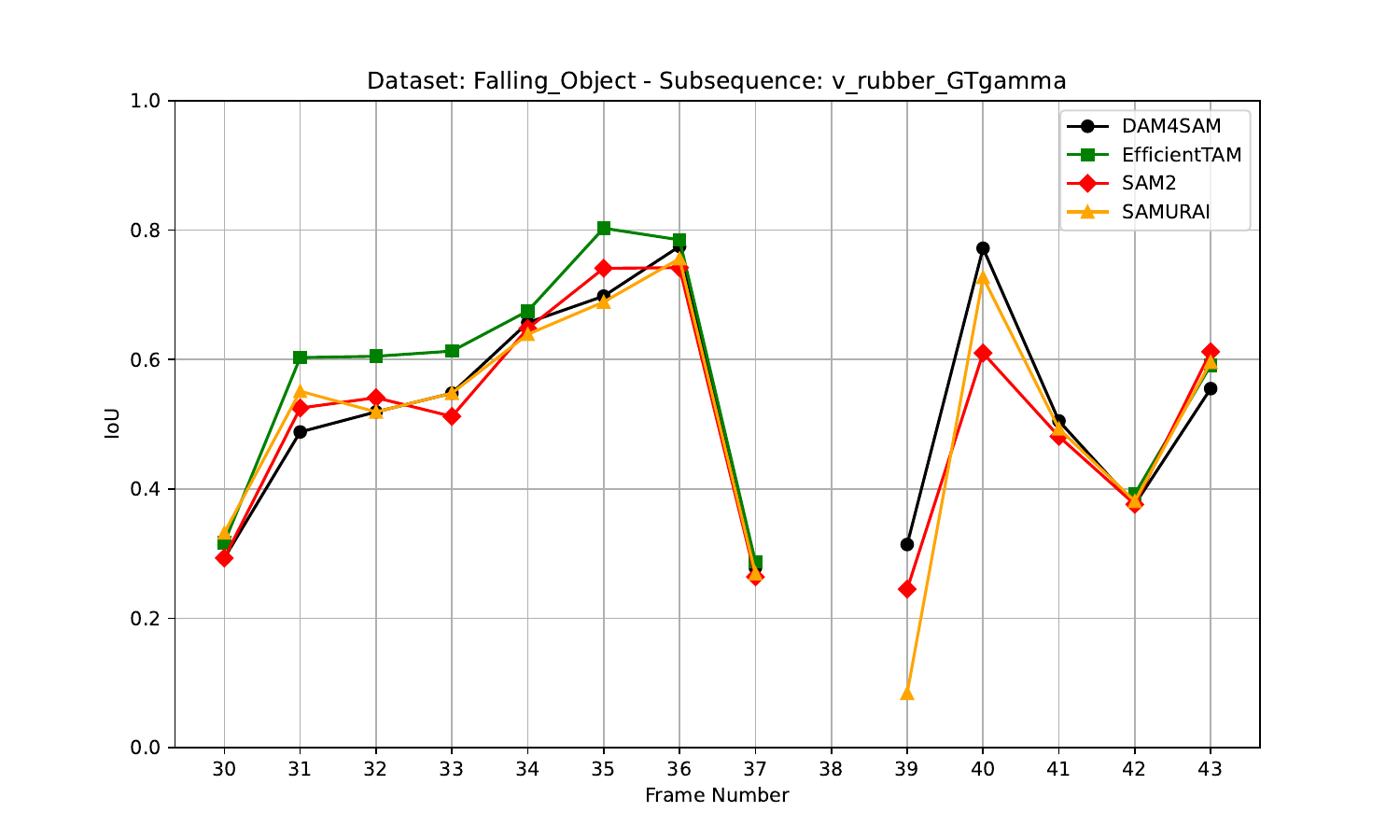}\\
\includegraphics[width=\textwidth]{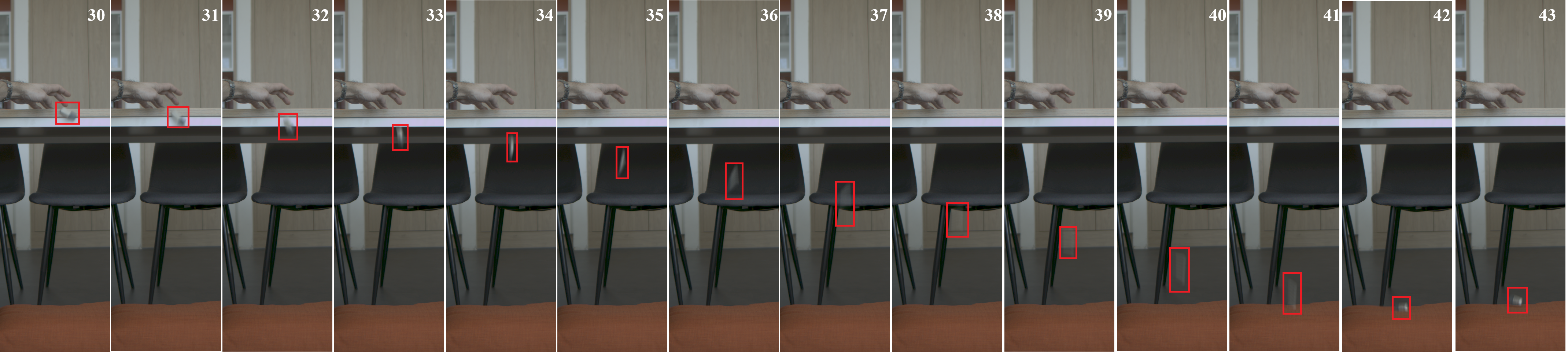}\\
\end{tabular}
\caption{Tracking performance (IoU) across frames on  the sequence \texttt{v\_rubber\_GTgamma} from dataset \textit{Falling Object}. All trackers fails to propose a bounding box for frame 38 while EfficientTAM also fails for frames 39 to 41 included. Corresponding frame numbers are given on top of each frame, and object (rubber) locations are indicated with red ground truth bounding boxes.}
\label{fig:frame_based_graph1}
\end{figure}

\begin{figure}[!h]
\begin{tabular}{c}
\includegraphics[width=\linewidth,trim=1cm .5cm 1.5cm 1cm,clip]{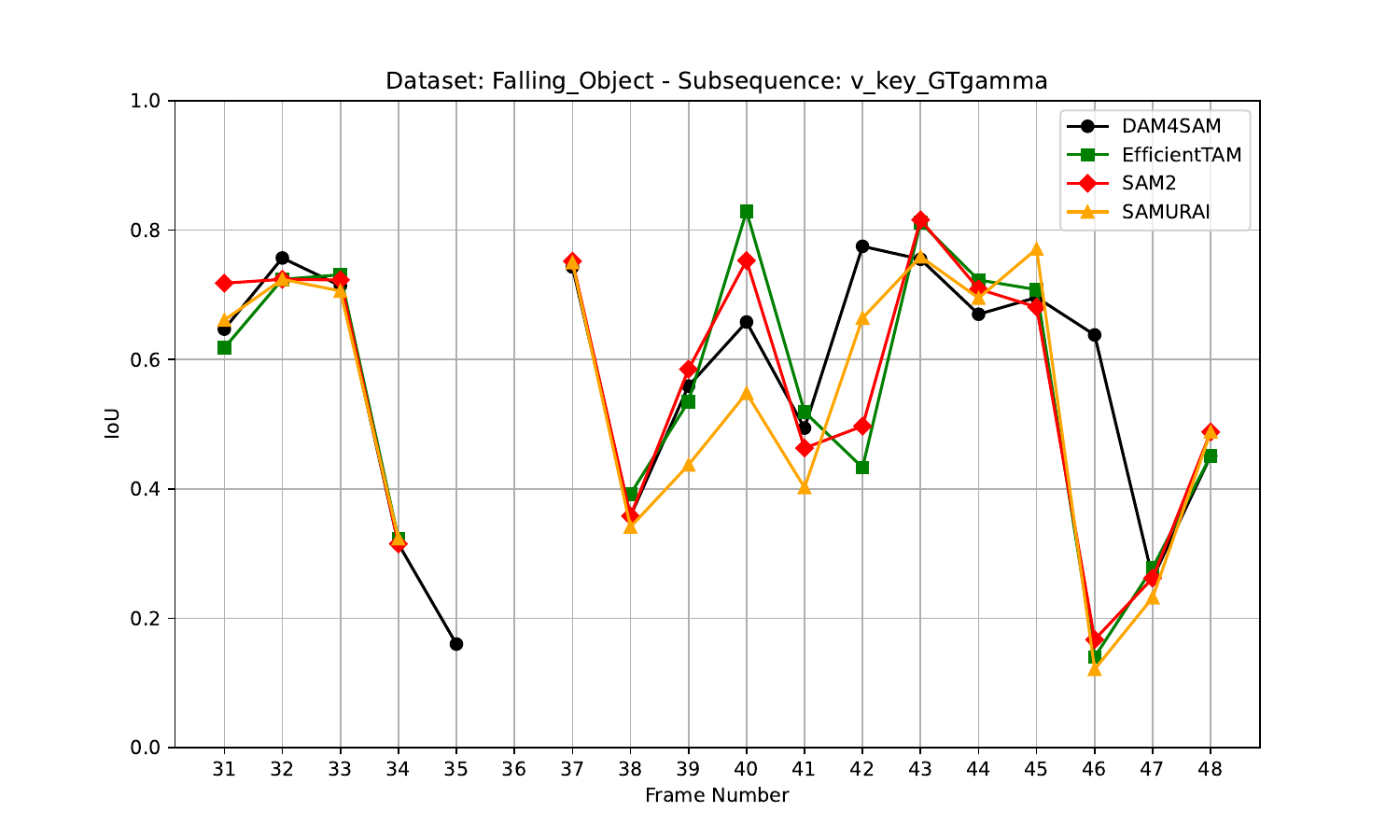}\\
\includegraphics[width=\textwidth]{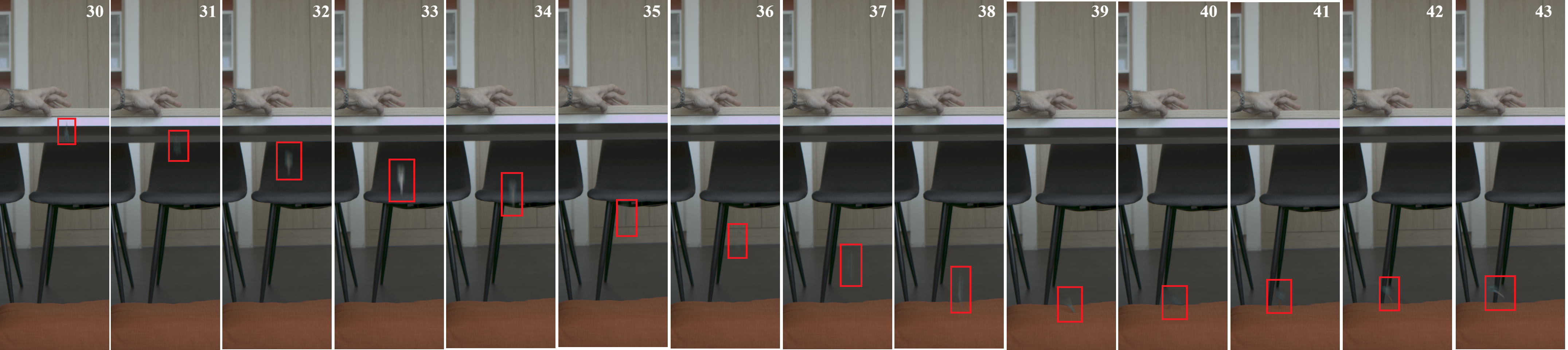}\\
\end{tabular}
\caption{Tracking performance (IoU) across frames on the sequence \texttt{v\_key\_GTgamma}  from \textit{Falling Object} dataset. All trackers fail to propose a bounding box for frame 36. For frame 35, DAM4SAM is the sole successful tracker.} 
\label{fig:frame_based_graph2}
\end{figure}

\begin{figure}[!h]
\begin{center}
\begin{tabular}{c}
\includegraphics[width=\linewidth,trim=1cm .5cm 1.5cm 1cm,clip]{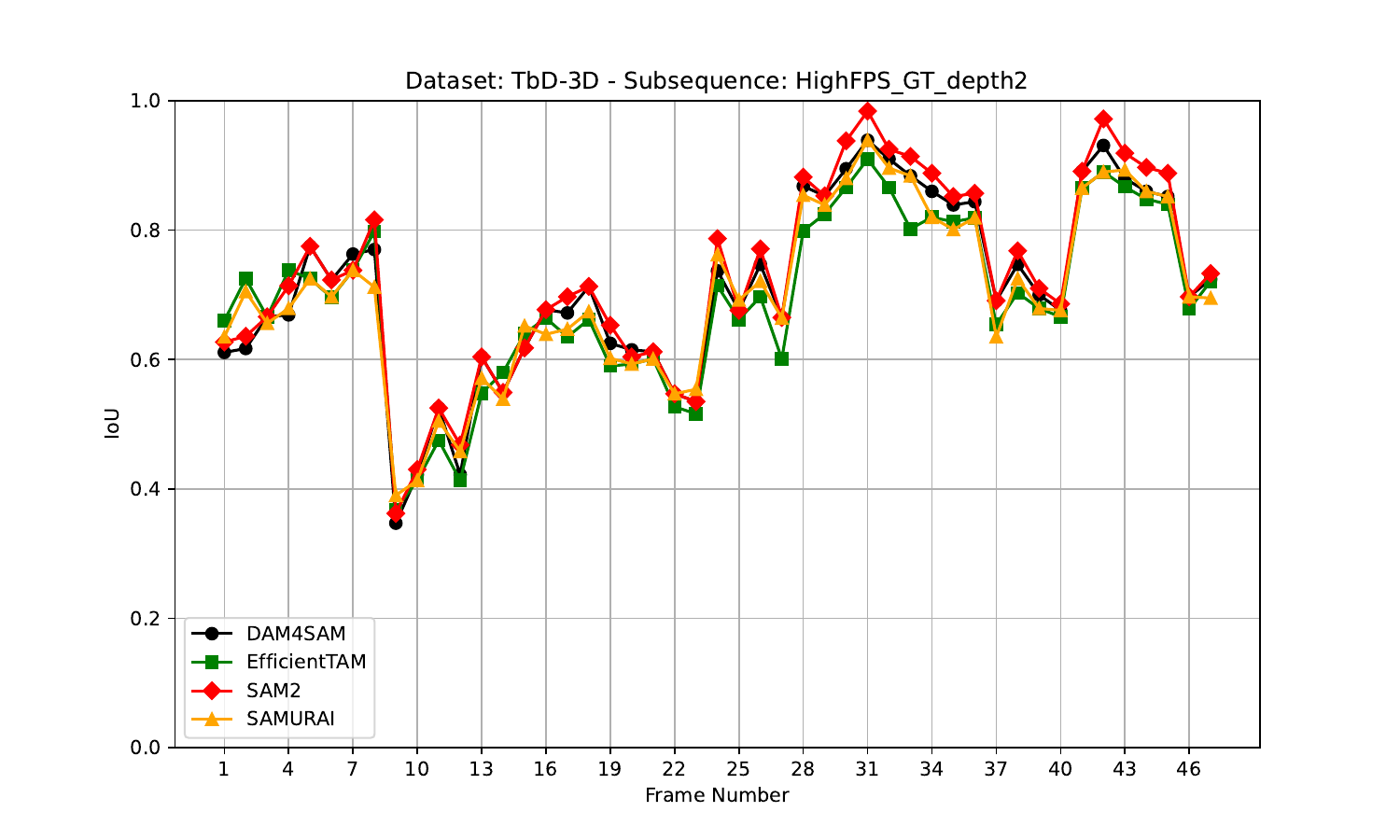}\\
\includegraphics[width=\linewidth,trim=1cm .5cm 1.5cm 1cm,clip]{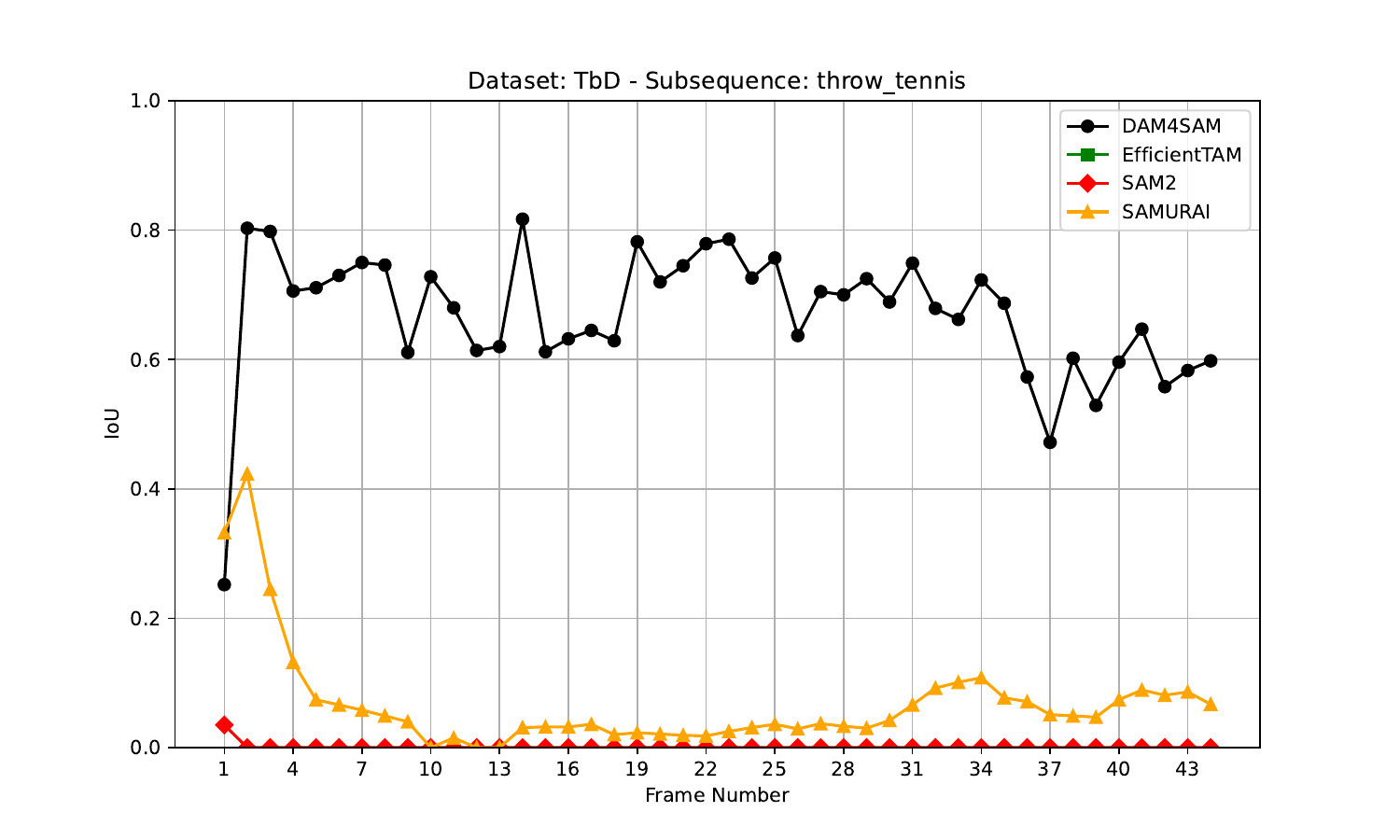}
\end{tabular}
\end{center}
\caption{Examples of tracking performance (IoU) across Frames in sequences \texttt{HighFPS\_GT\_depth2} in \textit{TbD-3D} dataset (top: all trackers perform well and provided similar results) and  \texttt{throw\_tennis} from \textit{TbD} dataset (bottom: all trackers performed poorly, with the exception of DAM4SAM; EfficientTAM failed to initialize for tracking  due to the strong motion blur present on the object, resulting in no performance curve being generated for this sequence in the graph). } 
\label{fig:frame_based_graph3}
\end{figure}

To better understand the temporal dynamics of each trackers performance, we analyze the frame-by-frame IoU plots as shown in Figures \ref{fig:frame_based_graph1}, \ref{fig:frame_based_graph2}, and \ref{fig:frame_based_graph3}. These figures highlight the behavior and failure points of each tracker throughout the sequences. The x-axis represents frame numbers derived from the sequence indicated in the frame names.
Trackers occasionally fail to detect objects when they become nearly invisible for one or two frames, most likely due to motion blur. For instance, for the sequence \texttt{v\_rubber\_GTgamma} in the \textit{Falling Object} dataset,  the EfficientTAM tracker exhibited multiple missed detections  (see Fig. \ref{fig:frame_based_graph1}). Notably, it failed to produce a prediction for frames 39, 40, and 41, immediately following a frame with strong motion blur (frame 38). While other trackers show similar transient failures, EfficientTAM is more consistently vulnerable to this prolonged loss of tracking following motion blur events. 
Similarly, this behavior is observed in Figure \ref{fig:frame_based_graph2}. While other trackers recovered after just one or two frames of failure, EfficientTAM  exhibited a more prolonged failure, missing the object for three consecutive frames (35, 36, and 37). 
On the other hand, in some sequences, the correlations between the trackers' performances are very strong (see top plot  Fig. \ref{fig:frame_based_graph3}). Conversely, in other sequences,  one or a few trackers perform exceptionally well while others perform near zero (see bottom plot  Fig. \ref{fig:frame_based_graph3}).

\section{Conclusion} \label{section:conclusion}

We have benchmarked several trackers on several datasets with fast moving objects, and we have shown that  both SAMURAI and DAM4SAM trackers outperform SAM2 and EfficientTAM.
Using FMOX classification of object sizes \cite{Aktas2025} for these datasets, we note that datasets presenting  sequences with smaller moving objects (and in addition with non overlapping ground truth bounding boxes between successive frames) affect tracker performance as measured by mIoU and mDice.

\begin{credits}
\subsubsection{\ackname} This research was supported by funding through the  Maynooth University Hume Doctoral Awards. For the purpose of Open Access, the author has applied a CC BY public copyright license to any Author Accepted Manuscript version arising from this submission.

\subsubsection{\discintname}
The authors declare they have no competing interests.
\end{credits}

%

\end{document}